\documentclass[preprint,12pt]{elsarticle}

\usepackage{amsmath,amsfonts,bm}

\def\eqref#1{equation~\ref{#1}}

\def\floor#1{\lfloor #1 \rfloor}
\def\1{\bm{1}}

\def\vg{{\bm{g}}}
\def\vh{{\bm{h}}}

\def\vu{{\bm{u}}}

\def\mE{{\bm{E}}}

\def\mG{{\bm{G}}}

\def\mK{{\bm{K}}}

\def\mQ{{\bm{Q}}}

\def\mV{{\bm{V}}}
\def\mW{{\bm{W}}}

\DeclareMathAlphabet{\mathsfit}{\encodingdefault}{\sfdefault}{m}{sl}
\SetMathAlphabet{\mathsfit}{bold}{\encodingdefault}{\sfdefault}{bx}{n}

\newcommand{\R}{\mathbb{R}}

\newcommand{\softmax}{\mathrm{softmax}}

\usepackage[utf8]{inputenc} 
\usepackage[T1]{fontenc}    
\usepackage[hidelinks]{hyperref}       
\usepackage{url}            
\usepackage{booktabs}       
\usepackage{amsfonts}       
\usepackage{nicefrac}       
\usepackage{microtype}      
\usepackage{xcolor}         
\usepackage{multirow}
\usepackage{graphicx}
\usepackage{caption}
\usepackage{subcaption}
\usepackage{bbold} 
\usepackage{wrapfig}
\usepackage{amsthm}
\usepackage{mathtools}
\usepackage{censor}
\usepackage{enumitem}
\usepackage{wrapfig}
\usepackage{float}
\usepackage{array}
\usepackage{adjustbox}

\journal{Signal Processing}

\begin{document}

\begin{frontmatter}
	\title{Asynchronous Graph Generator}
	
	\author[1]{Christopher P. Ley\corref{cor1}}
	\ead{info@christopherley.com}
	\cortext[cor1]{Corresponding Author}
	\affiliation[1]{organization={Research \& Development, Machina Doctrina},
	city={Santiago},
	country={Chile}}

	\author[2]{Felipe Tobar}
	\ead{f.tobar@imperial.ac.uk}
	\affiliation[2]{organization={Department of Mathematics \& I-X, Imperial College},
	city={London},
	country={United Kingdom}}

	\begin{abstract}
		We introduce the asynchronous graph generator (AGG), a novel graph attention network for imputation and prediction of multi-channel time series. Free from recurrent components or assumptions about temporal/spatial regularity, AGG encodes measurements, timestamps and channel-specific features directly in the nodes via learnable embeddings. Through an attention mechanism, these embeddings allow for discovering expressive relationships among the variables of interest in the form of a homogeneous graph. Once trained, AGG performs imputation by \emph{conditional attention generation}, i.e., by creating a new node conditioned on given timestamps and channel specification. The proposed AGG is compared to related methods in the literature and its performance is analysed from a data augmentation perspective. Our experiments reveal that AGG achieved state-of-the-art results in time series imputation, classification and prediction for the benchmark datasets \emph{Beijing Air Quality}, \emph{PhysioNet ICU 2012} and \emph{UCI localisation}, outperforming other recent attention-based networks.
	\end{abstract}

%
%
%
%
	
\end{frontmatter}
%
	
	\section{Introduction}
	Incomplete time series data are ubiquitous in a number of applications \cite{miao2019answering}, including medical logs, meteorology records, traffic monitoring, financial transactions and IoT sensing. Missing records may be due to various reasons which include failures either in the acquisition or transmission systems, privacy protocols, or simply because the data are collected asynchronously in time. Missing data is an issue in itself but also hinders applications, for example, the public dataset PhysioNet \cite{silva2012predicting} has a 78\% average missing rate which makes it challenging to extract useful information for, e.g., for predicting mortality. In this setting, imputation refers to filling in the missing values using the available observations \cite{little2019statistical}, and can be achieved by methods that exploit both temporal and spatial dependencies \cite{yoon2017multi,yi2016st}.
	\\
	
	Existing approaches \cite{cao2018brits, cini2021filling} to imputation in multi-channel time series often assume temporal regularity of the data, which is a consequence of representing the values of the series through a matrix with missing entries. This representation implicitly induces two critical biases: i) the notion of order (causality), e.g., $x_1$ precedes $x_2$, and ii) a fixed sampling rate implying synchronous data acquisition. We assert that this representation is detrimental to successfully learn the latent dynamics generating the (sparse) observations, therefore, we propose to relax these stringent assumptions and represent observations (samples/measurements) as nodes in an asynchronous graph, as depicted in Fig.~\ref{fig:data_collection}. This approach is robust to the occurrence of missing data and exploits the permutation invariance of multiple channels of sensors. It also enables imputation as transductive node generation over graph embeddings through a novel mechanism we refer to as \emph{conditional attention generation} (see Sec~\ref{sec:conditional_attention}). 
	\\
	
	Deep learning approaches to data imputation have become increasingly popular in the last five years \cite{yoon2018estimating,yoon2018gain,liu2019naomi,cao2018brits,cini2021filling,yildiz2022multivariate}. However, in general these methods rely on slight modifications of standard neural architectures tailored for complete discrete-time data and are thus unable to fully discover relationships among temporal or channel-specific features \cite{bai2018empirical,chung2014empirical}. We argue that dropping assumptions about temporal regularity (or \emph{synchronicity}) would allow for graph architectures to discover rich relationships across the channels of the time series. Therefore, we assume no data regularity beyond what is explicitly observed through each sensor, all with the aim to learn the latent dynamics as agnostically as possible and then perform imputation by generating new nodes. Consequently, our approach is termed \emph{Asynchronous Graph Generator} (AGG).\\
	\\
	The main contributions of our work are:
	\setlist{nolistsep}
	\begin{itemize}[noitemsep, leftmargin=10pt]
		\item An interpretation of multi-channel time series as an asynchronous graph, where nodes represent the observations.
		\item Novel use of embeddings to encode temporal and channel features which can be leveraged through attention to model the relationships among observations.
		\item The introduction of \emph{conditional attention generation} (see Sec.~\ref{sec:conditional_attention}), a mechanism to generate new observations conditioned on given temporal/channel features.
		\item An experimental validation of AGG on standard benchmarks against the state of the art.
		\item A study of the limiting performance of AGG from the perspective of data augmentation.
	\end{itemize}
	
	\section{Related work}\label{sec:related_work}
	\noindent
	A number of deep learning models have been successfully developed for multi-channel time-series imputation, in particular, using recurrent neural networks (RNNs) \cite{cao2018brits,yoon2018estimating,lipton2016directly,che2018recurrent,luo2018multivariate}. Notably, GRU-D \cite{che2018recurrent} analyses sequences with missing data by controlling the decay of the hidden states of a gated RNN, while BRITS \cite{cao2018brits} implements a bidirectional GRU-D that incorporates cross-channel correlation to perform spatial imputation. These RNN-based methods assume temporal regularity of data, i.e., a fixed sampling (acquisition) rate.\\

	{
	Adversarial strategies have also been applied to imputation. For instance, GAIN \cite{yoon2018gain} uses GANs \cite{goodfellow2020generative} to perform imputation in the i.i.d.~setting thus ignoring dependencies among sensors, while \cite{luo2018multivariate,luo2019e2gan} train models to generate realistic synthetic sequences. An approach similar to GAIN, termed SSGAN \cite{miao2021generative}, implements a generator conditioned on the predicted label to reconstruct  missing values. Lastly, NAOMI \cite{liu2019naomi} addressed the imputation problem for multi-scale highly-sparse series using hierarchical models.}\\

	{Concurrently, graph neural networks (GNN) \cite{9046288} have found applications in spatio-temporal forecasting, where the idea underpinning most methods is the extension of RNN architectures to the graph domain. For instance, \cite{seo2018structured} implemented GRU cells as nodes combined with spectral GNN operations \cite{defferrard2016convolutional,cini2021filling} similarly use RNNs to exploit temporal relationships and utilise message passing GNNs to exploit spatial (topological) graphs, while \cite{li2018diffusion} replaced spectral GNNs with a diffusion-convolutional network \cite{atwood2016diffusion}. Spatio-temporal graph convolutional networks that alternate convolutions on temporal signals and spatial signals have been proposed by \cite{scarselli2008graph, li2015gated, yu2017spatio, wu2019graph, wu2020connecting}. 
	}\\
	
	Recently, Transformers \cite{vaswani2017attention} have achieved impressive results on imputation tasks. Combining Transformer-like architectures with RNNs, \cite{cai2020traffic,zhang2018gaan} focused on spatio-temporal data, while \cite{yildiz2022multivariate} leverages a Transformer based auto-encoder with a learnable temporal encoding, and a masking scheme similar to that proposed under the BRITS architecture \cite{cao2018brits}. Stemming from the observation that transformers can be framed as special case \cite{velivckovic2023everything} of Graph Attention Neural Networks \cite{brodyattentive,velivckovicgraph}, \cite{yildiz2022multivariate} proposed a Transformer with temporal embeddings which indirectly produces an asynchronous graph of multi-channel time series, where the transformer learns a graph representation via the dot product self-attention over tokens with flexible positioning. We argue that this approach is promising but the architecture in \cite{yildiz2022multivariate} does not go far enough as it decouples the signal from channel information; thus, we propose the Asynchronous Graph Generator (AGG) to exploit both channel and temporal features through attention. In the AGG, predictions are computed a novel transductive operation termed \emph{conditional attention generation}, which aims to avoid limitations identified in other architectures, such as autoencoders, when it comes to prediction (i.e., new sample generation).
	\\
	
	To the best of our knowledge, no previous attention based (graph or transformer) method approaches the imputation problem from the perspective of an asynchronous graph implementing transductive node generation.
	
	\section{The AGG architecture}
	\label{sec:architecture}
	
	\begin{figure}[t]
		\centering
		\includegraphics[width=1.05\linewidth]{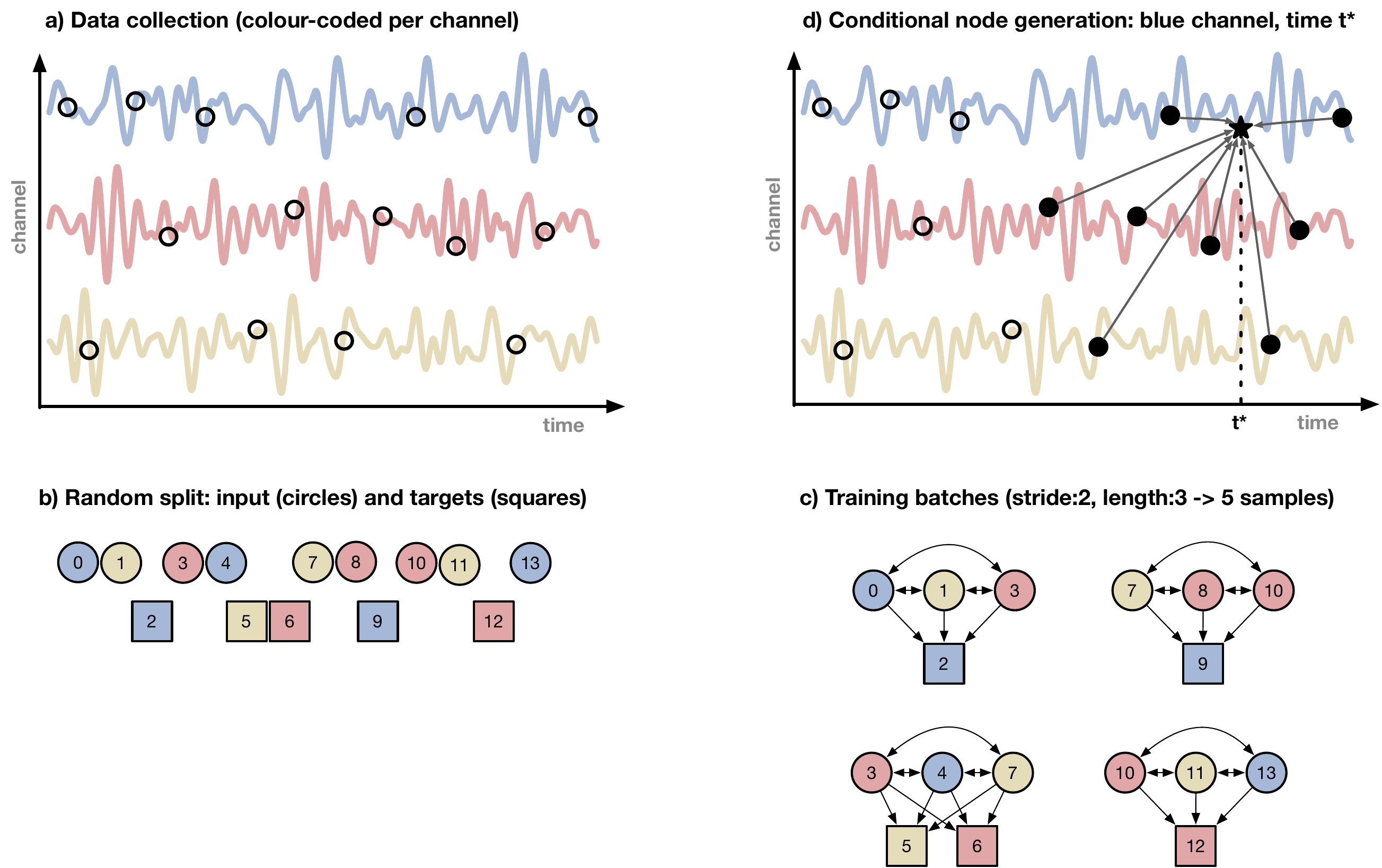}
		\caption{An illustration of the AGG self-supervised pipeline. \textbf{a)} Time series samples are collected (possibly) asynchronously, and comprise measurements, timestamps and channel features. \textbf{b)} Samples are ordered and a split into inputs and targets for self-supervised training. \textbf{c)} The input/target split is considered as instances of the asynchronous graph for training. \textbf{d)} The learnt graph encodes a rich representation of the underlying signal, where new samples to be generated through \emph{conditional attention generation}; here, $c_g = blue$ and $\tau_g = t_N - t_*$.}
		\label{fig:data_collection}
	\end{figure}

	{The rationale behind AGG is to model each observation as a node within a homogeneous graph and then learn the dependencies (the graph) between all the nodes by leveraging the attention mechanism. Estimations of (unseen) observations can be performed by generating a new node in the graph, the value of which is calculated via the learnt node relationships}. This means that AGG is a homogeneous dynamic graph admitting node additions \cite{rossi2020temporal}. The assumption of homogeneity is crucial, since it ensures that all the required information to generate a new node is encoded in the neighbouring nodes and not provided by region-specific edges. Specifically, imputation occurs via \emph{conditional attention generation} (see Sec:~\ref{sec:conditional_attention}), that is, if nodes are composed by measurements, timestamps, and channel-specific features (e.g., location or type), AGG generates the measurement corresponding to a given timestamp and channel features. We emphasise the timestamps need not be uniformly sampled or even ordered. 
	\\
	
	Though the main application of the AGG in this paper is data imputation, the expressive representation of the graph embeddings allows for addressing other downstream tasks such as prediction, classification, compression, and anomaly detection. This motivates the interpretation of data imputation as self-supervised pre-training through masked data augmentation \cite{Balestriero2022}, similarly to what architectures like BERT \cite{devlin2018bert} do in NLP applications.
	
	\subsection{Problem formulation and model overview}
	\label{sec:data_prep}
	Let us consider a set of sensor observations $\mathcal{D}=\{x_n\}_{n=1}^N$. The $n$-th observation is given by
	\begin{equation}
		\label{def:node}
		x_n= [y_n, t_n, c_n]\in \R^{d_y+1+ d_c},
	\end{equation}
	where $y_n\in \R^{d_y}$ is the \textbf{measurement}, $t_n\in \R$ is the \textbf{timestamp} and $c_n\in \R^{d_c}$ is a collection of \textbf{channel} features including --but not limited to-- location, operating conditions of the measurement, type (e.g., pressure, temperature, speed), and any other feature specific to the channel. Our aim is to leverage on  $\mathcal D$ to predict \textbf{measurements} corresponding to a set of new \textbf{timestamps} and \textbf{channels}. To exemplify the role of this notation consider the Beijing dataset, where \textbf{channel} captures the measurements' type (e.g., PM2.5, pressure, temperature) as well as their location. Our formulation stems from the assumption that measurements across the graph are related not only by their timestamps but also by the sensor's type and location. Explicitly encoding these features (timestamp and channel) in the nodes allows the graph to be learnt in a way that exploits the interactions among the relevant variables, e.g., sensors of different types should interact differently as should different physical locations. We conjecture that such encodings allow the graph to be fully context aware and thus performant.
	\\
	
	In a nutshell, AGG is implemented as follows. First, the training set $\mathcal D$ in \eqref{def:node} is obtained via an acquisition system (Fig.~\ref{fig:data_collection}a) and each measurement is considered as a node in a graph. Second, the nodes are sorted wrt their timestamps and randomly split into input and target samples (Fig.~\ref{fig:data_collection}b). Third, the dataset is divided into samples of $L$ inputs and a single output by sequentially passing through the observations with a stride $\Delta$ (Fig.~\ref{fig:data_collection}c); here, the AGG is trained (see Appendix~\ref{app:block_setup} for more detailed instructions). Fourth, for a given timestamp $t^\star$ and channel features, prediction is performed by generating a new node (Fig.~\ref{fig:data_collection}d).
	
	\subsection{Learnable embeddings for measurements, timestamps and channels}
	
	\textbf{Temporal embedding.} Graphs are naturally permutation invariant so in order to learn flexible representations of temporal differences, such as periodicities and long-range dynamics, we must encode the temporal position along with nodes features. Following \cite{kazemi2019time2vec}, we  use the learnable temporal encoding $\mathbf{t2v}$ and then use these learnt representation in a similar vein as positional encoding in \cite{vaswani2017attention}.  For an observation  $x_n$ as defined in \eqref{def:node}, this embedding is parametrised as
	\begin{align}\label{eq:time2vec}
		\mathbf{t2v}(\tau_n)= [\;\omega_0\tau + \varphi_0,\; \mathcal{F}\left(\omega_1\tau_n + \varphi_1\right),\; \ldots,\; \mathcal{F}\left(\omega_{D_t-1}\tau_n + \varphi_{D_t-1}\right)\; ]^\top \in\R^{D_t},
	\end{align}
	where $\tau_n$ is the temporal difference between $x_n$ and last-observed node $x_N$, i.e., $\tau_n = t_N - t_n\geq0$; $\{\omega_k\}_k$ and $\{\varphi_k\}_k$ are learnable parameters; and $\mathcal{F}$ is a periodic function. Inspired by \cite{kazemi2019time2vec}, we choose $\mathcal{F}(\cdot) = \sin{(\cdot)}$ in all implementations of AGG.
	\\
	
	\textbf{Channel embedding.} In order to utilise measurements of different nature (defined by the channel-specific features) one could be tempted to represent all interactions via a heterogeneous graph and build specific models for each interaction of nodes and edges. However, this would require us to cater for all possible relationships among nodes with minimal weight sharing throughout the model. To circumvent this challenge, AGG is modelled as a homogeneous graph instead, where a single learnable form of interaction operates over values $y_n$, timestamps $t_n$ and channel features $c_n$ provided by the sensor measurement. In the same vein as the temporal embedding, the channel is represented by a set of learnable embeddings, a practice that has become prevalent in the field of natural language processing for \emph{learnable word embeddings} beginning with \cite{bengio2000neural}. This way, we aim to include all available information as a form of inductive bias \cite{bronstein2021geometric} into the model, and leave the graph structure to exploit rich relationships among features and values via an attention mechanism.
	\\
	
	AGG constructs channel embeddings based on whether they are discrete or continuous: discrete channel features (e.g., categorical data) are embedded via hashing, that is, a matrix of learnable weights is sliced at the index of the relevant category. Similarly, continuous channel features are embedded into higher dimensions through a learnable projection matrix. The complete channel embedding (discrete and continuous) is denoted $\text{embed}(c_n)\in\R^{D_c}$.
	\\
	
	To enhance the representation power of the overall architecture, we follow \cite{velivckovicgraph} and also include a learnable projection for the measurement value denoted $\text{embed}(y_n)\in \R^{D_y}$. Therefore, the AGG is a homogeneous graph $\mathcal{G}$ with $n$-th node containing  
	\begin{equation}
		\label{eq:h0}
		h_0 = \text{Concat}\left[\text{embed}(y_n), \mathbf{t2v}(\tau_n), \text{embed}(c_n)\right]\in \R^{D_y+D_t+D_c},
	\end{equation}
	where the explicit dependence on the index $n$ is ignored unless necessary.
	\\
	
	Observe that we denoted the original dimensions in lowercase ($d_y$ and $d_c$) and the embedded ones in uppercase ($D_y$, $D_t$ and $D_c$). Also, following \eqref{eq:h0} we define the encoder dimension  $d_\text{encoder} = \text{dim}(h_0) = D_y+D_t+D_c$, where the notation $h_0$ will be clarified in the next section.  
	\\
	
\begin{figure*}[ht]
		\centering
		\includegraphics[width=\linewidth]{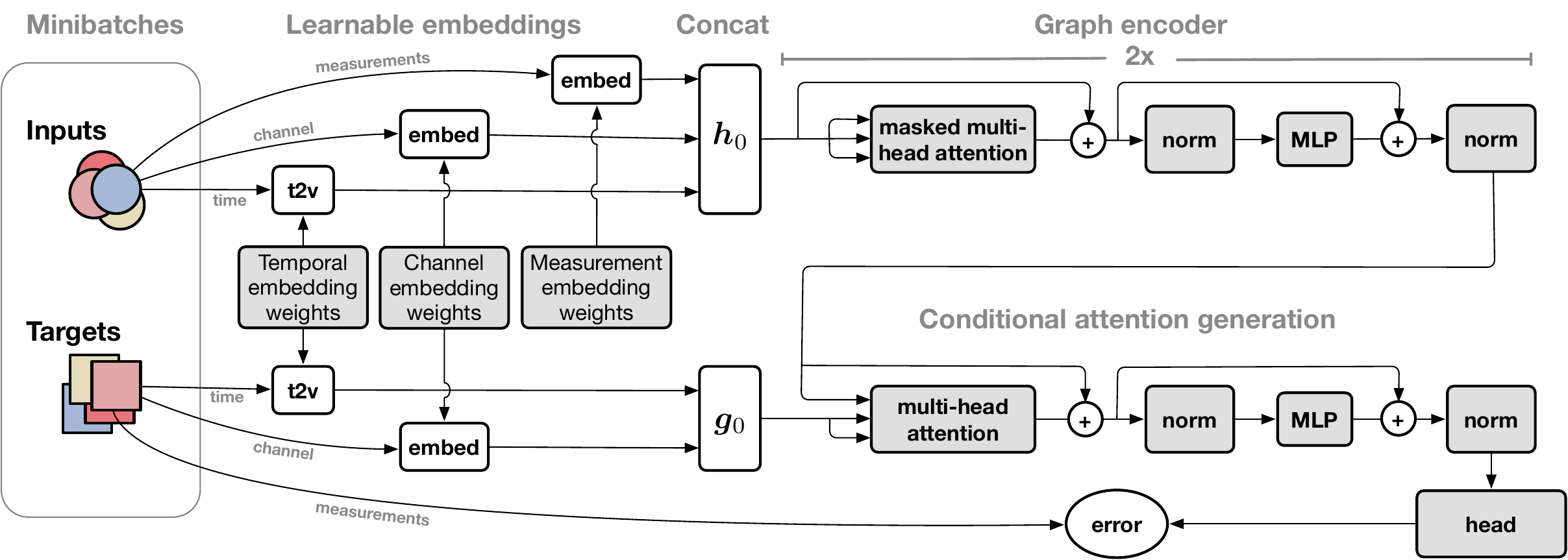}
		\caption{AGG architecture: The sections of the network are indicated at the top of the figure. Inputs and target are represented as circles and squares respectively, fixed operations are denoted by white blocks and learnable transformations in light grey blocks.}
		\label{fig:architecture}
	\end{figure*}

	Fig.~\ref{fig:architecture} illustrates the embedding procedure under the title \textbf{learnable embeddings}. The  embeddings then enter a sequence of encoder and decoder (generator) blocks comprising attention and fully-connected layers with layer-norms and skip connections through addition. The next two sections present the encoder and the generator stages. 
	
	\subsection{Asynchronous graph encoding}
	\label{sec:graph_encoder}
	Towards improved performance and expressibility \cite{brodyattentive,velivckovicgraph, vaswani2017attention}, the encoder features a multi-head self-attention layer representing the interactions among values, timestamps, and channel features.
	\\
	
	Following \eqref{eq:h0}, for a single node we denote $h_{i-1}$ and $h_i$ the input and output of the $i$-th encoder block respectively ($i\geq1$). However, recall from Sec.~\ref{sec:data_prep} that AGG takes $L$ nodes simultaneously, thus, we denote $\vh_i$ as the concatenation of the $h_i$'s coming from these $L$ nodes. Therefore, each $\vh_i \in \R^{L \times d_{\text{encoder}}}$ is a tensor comprising $L$ node embeddings.
	\\
	
	The $j$-th head of the $i$-th attention layer is thus given by \cite{vaswani2017attention}:
	\begin{align}
		\label{eq:attention-layer}
		\text{Attention}(\mQ_{ij}, \mK_{ij}, \mV_{ij}) &= \softmax{\left(\mE_{ij}\right)}\mV_{ij} \in\R^{L \times d_v},
	\end{align}
	where:
	
	\begin{itemize}[leftmargin=10pt]
		\item $\mQ_{ij} = \vh_{i-1}\mW^Q_j\in \R^{L \times d_q}$, $\mK_{ij} = \vh_{i-1}\mW^K_j\in \R^{L \times d_k}$, $\mV_{ij} = \vh_{i-1}\mW^V_j\in \R^{L \times d_v}$ are the query, key and value embeddings respectively.
		\item $\mW^Q_i \in \R^{d_{\text{encoder}}\times d_q}$, $\mW^K_i \in \R^{d_{\text{encoder}}\times d_k}$, $\mW^V_i \in \R^{d_{\text{encoder}}\times d_v}$ are the projection matrices.
		\item $\mE_{ij} = d_k^{1/2}\mQ_{ij}\mK_{ij}^\top\in\R^{L\times L}$ is the dot product attention \cite{vaswani2017attention} matrix, which is equivalent to a fully-connected weighted graph \cite{velivckovic2023everything}. Under the graph interpretation, $\mE$ is the weighted adjacency matrix for the $L$ nodes in the asynchronous graph, where the weight represents the relevance of neighbouring nodes in determining the features of any other node. 
		
	\end{itemize}
	\vspace*{10px}
	Then, the $i$-th multihead attention layer is simply the weighted concatenation of its $l$ attention heads:
	\begin{align}
		\text{MultiHead}_i &= \text{Concat} [\text{Attention}(\mQ_{i1}, \mK_{i1}, \mV_{i1}),\ldots,\nonumber\\ 
		&\text{Attention}(\mQ_{il}, \mK_{il}, \mV_{il})]\mW^O \in\R^{ld_v\times d_{\text{encoder}}}.
		\label{eq:multihead}
	\end{align}
	
	Lastly, the output of the $i$-th multi-head attention is normalised via a layer normalisation \cite{ba2016layer} followed by a multi-layer perceptron (MLP). The MLP consists of a 2-layer feed-forward network with a LeakyReLU \cite{maas2013rectifier} activation and Dropout \cite{hinton2012improving} in the hidden layer, followed by a linear activation layer. The MLP has layer sizes of $[d_{\text{encoder}}, l\times d_\text{encoder}, d_\text{encoder}]$, with $l$ the number of heads. Throughout each block, there is extensive use of skip connections following inspiration from the Transformer \cite{vaswani2017attention} and the original introduction of the residual connections, ResNet \cite{he2016deep}.
	
	The output of the $i$-th block is then calculated by:
	\begin{align}
		\vu_i &= \vh_{i-1} + \text{MultiHead}_{i} \label{eq:encoder_in}\\
		\vh_i &= \text{LayerNorm}\left[\vu_i + 	\text{MLP}\left(\text{LayerNorm}\left[\vu_i\right]\right)\right] \label{eq:encoder_out}.
	\end{align}
	Therefore, Eqs.~\ref{eq:attention-layer}~-~\ref{eq:encoder_out} define the outputs from the asynchronous graph encoder blocks $\vh_0,\ldots,\vh_l$.
	
	\subsection{Conditional attention generation}\label{sec:conditional_attention} 
	{Predictions in the proposed model are performed by creating a new node in the graph with given input features. Then, as the graph operates in a homogeneous manner, the value of the new node will be computed from its relationship with neighbouring nodes through an operation termed \emph{conditional attention generation} described in this section. Intuitively, the prediction of the measurement in the new node is computed via an cross-attention network (see Fig.~\ref{fig:architecture}) between: i) information about the time, channel and measurement of the neighbouring nodes in the form of the output of a self-attention network denoted $\vh_l$, and ii) the conditioning information of the new node comprising only time and channel features denoted $\vg$. The choice of a cross-attention mechanism is motivated by its proven ability to find complex relationships among variables of different types.}

	Specifically, the conditioning variable is given by
	\begin{equation}
		\vg = \text{Concat}[\mathbf{t2v}(\tau_g), \text{embed}(c_g))] \in \R^{d_g},
	\end{equation}
	where $d_g = D_c + D_t$, and again we reiterate that $\tau_g$ and $c_g $ are user defined.  The relative importance of the node embeddings to the conditioning vector $\vg$ is calculated by the following attention mechanism:
	\begin{align*}
		\bar{\vg} &= \text{LayerNorm}[\vg],\nonumber\\
		\mG &= \bar{\vg}\overline{\mW}^G\in\R^{d_{\bar{q}}},\nonumber\\ 
		\mK_{j} &= \bar{\vh}_l\overline{\mW}^K_j\in \R^{L\times d_k},\nonumber\\ 
		\mV_{j} &= \bar{\vh}_l\overline{\mW}^V_j\in\R^{L\times d_v} 
	\end{align*}
	\begin{equation}
		\text{ConditionalAttention}(\mG, \mK_{j}, \mV_{j}) = \softmax{\left(d_k^{1/2}\mG\mK_{j}^T\right)}\mV_{j} \in\R^{d_v}.\label{eq:conditional attention}
	\end{equation}
	Conditional attention generation also implements multiple attention heads ($h$), which, akin to Eqs.~\ref{eq:attention-layer}~\&~\ref{eq:multihead}, are
	\begin{align}
		\text{CondMultiHead}&= \text{Concat}[\text{Attention}(\mG_{1}, \mK_{1}, \mV_{1}),\ldots,\nonumber\\
		&\text{Attention}(\mG_{l}, \mK_{l}, \mV_{l})]\mW^O\in\R^{d_y},\label{eq:cross-multihead}\\
		\text{where}~&\overline{\mW}^G\in\R^{d_g \times d_{\bar{q}}}, ~\overline{\mW}^K_j\in \R^{d_{\text{encoder}}\times d_k}\nonumber\\
		&\overline{\mW}^V_j\in\R^{d_{\text{encoder}}\times d_v},~\mW^O\in\R^{hd_v\times d_y}\nonumber
	\end{align}
	are the  projection matrices. Additionally, similar to the asynchronous encoder block, the generator follows the cross attention layer with a set of LayerNorms, skip connection and an MLP, such that:
	\begin{align}
		\bar\vu &= \text{CrossMultiHead} \label{eq:cross_in}\\
		\hat y_{n} &= \text{LayerNorm}\left[\bar\vu + 	\text{MLP}\left(\text{LayerNorm}\left[\bar\vu\right]\right)\right] \label{eq:cross_out},
	\end{align}
	where $\hat y_{n}$ is the estimated measurement value conditioned on the channel and time features $\tau_g$ and $c_g$ respectively. Fig.~\ref{fig:architecture} shows a diagram of the entire AGG architecture identifying the connections, inputs, targets, as well as fixed and trainable blocks.

	\section{Relationship to previous methods}
	

	The \emph{Asynchronous Graph Generator} (AGG) attempts to decouple the requirements of multi-channel time-series data from assumptions of regularity and exploit the dynamics of adjacent channels in a flexible way. It does so by representing each measurement as set of extended node features on a homogeneous graph which include conditioning information of both the relative time and channels from which they belong, this is a key component to encoding both the \emph{coherence} and \emph{phase} relationship \cite{granger1969investigating}, which quantify the similarity and delay between a pair of time series.
	\\
	
	Discovering the underlying graph relationship between nodes requires either directly encoding all or a subset of these features by hand \cite{cini2021filling} or include the graph definition as part of the learning process \cite{brodyattentive,velivckovic2023everything,velivckovicgraph, rossi2020temporal}. Recently, \cite{cini2021filling} proposed GRIN, which exploits the spatial graph relationship between channels by directly encoding its topological position as an adjacency matrix for a message passing graph neural network, where each node represents a source of a signal in a static graph. The temporal relationships of each measurements are then delegated, similarly to what was proposed by \cite{rossi2020temporal}, as a recurrent neural network. They implement a graph of bidirectional RNNs encapsulated in GNNs, where a series of RNNs are interconnected through gates controlled by message-passing NNs. These works consider the time series as a sequence of weighted directed graphs, thus assuming each node to be identified and labelled with a unique id and consistently available at all evenly-sampled timestamps. Therefore, their graphs have a fixed topology over time and thus the methods operate mainly by exploiting of network homophily.
	\\
	
	Similarly, AGG attempts to exploit the topological relationships between channels as a graph, but this knowledge instead is used as an inductive prior for the design. The AGG seeks to learn the graph relationship through the attention mechanism rather then considering a static one. This is achieved by retaining the acquired information about each channel as an embedding, which is learnt along with the rest of the model weights. This allows the notion of similarity among channels to be learnt alongside the representation of their interconnection. The core component modelling the relationship among channels is the attention mechanism, which compares channel embeddings to produce an adjacency matrix or a set of attention scores. This mechanism is mirrored with the temporal relationship through (temporal) embeddings and through the use of multi-head attention, where the AGG can learn multiple separate graphs representing channel-wise, temporal and feature relationships. These are then  combined to produce rich representations that can be exploited for various applications such as imputation through conditional attention generation.
	\\
	
	Transformer architectures with learnable position encoding, such as MTSIT \cite{yildiz2022multivariate}, similarly to the AGG, learn a flexible graph relationship of the temporal relationship of each measurement through the attention mechanism \cite{velivckovic2023everything}. Unlike the AGG, the MTSIT does not encode any channel-wise (e.g. topological) representation of the sources, i.e., all sources are considered equal. This, we argue, does not allow the exploitation of the latent dynamics between channels such as different phase relationships that depend on their topological distance. Lastly, MTSIT proposes an auto-encoded transformer so it is unclear how this could be used for upstream tasks such as prediction, unlike the AGG where the conditional generation can be applied directly to such tasks.
	\\
	
	As the two most similar architectures to the AGG are MTSIT \cite{yildiz2022multivariate} and GRIN \cite{cini2021filling} we focus on comparing the results of imputation directly to these two most-recent benchmarks. For further insight, we also compare AGG to other previous results such as SSGAN \cite{miao2021generative} and BRITS \cite{cao2018brits}, who similarly used masked imputation and performed a sensitivity analysis on additional relevant benchmarks; our experiments also include NAOMI \cite{liu2019naomi}, GP-VAE \cite{fortuin2020gp} for context.
	
	\section{Experimental evaluation}
	\label{sec:eval}
	\subsection{Preliminaries} 
	
	\noindent\textbf{Overview and experimental design.}  AGG was implemented, tested and compared to different state-of-the-art methods {through four experiments} described as follows.
	
	\begin{itemize}[leftmargin=10pt]
		\item \textbf{Experiment 1} (E1, Sec.~\ref{exp:missing}) assesses the performance of AGG of data imputation under different regimes of missing data against the following time-series models models: SSGAN \cite{miao2021generative}, BRITS \cite{cao2018brits}, NAOMI \cite{liu2019naomi}, GP-VAE \cite{fortuin2020gp}. 
		\item \textbf{Experiment 2} (E2, Sec.~\ref{exp:attention}) compares AGG against state-of-the-art methods that are close to our proposal in terms of their architecture: a Graph Neural Network (GRIN) \cite{cini2021filling} and a transformer-based method (MTSIT) \cite{yildiz2022multivariate}. 
		\item \textbf{Experiment 3} (E3, Sec.~\ref{exp:classification}) evaluates the flexibility of the graph embeddings produced by AGG in other downstream tasks (classification and prediction). Benchmarks in this case were BRITS \cite{cao2018brits} and GRIN \cite{cini2021filling}.
		\item {\textbf{Experiment 4} (E4, Sec.~\ref{exp:forex}) evaluates the AGG on annother application domain, which involves the interpolation of foreign exchange rate data under various levels of missing data.}
	\end{itemize} 
	
	\noindent\textbf{Datasets.} We considered three benchmark datasets: the \emph{Beijing Air Quality} \cite{yi2016st}, \emph{PhysioNet Challenge 2012} \cite{silva2012predicting} and \emph{UCI Localization Data for Person Activity} \cite{kaluvza2014multi} and one additional non-benchmark dataset: \emph{Foreign Exchange Rates}. Sec.~\ref{exp:missing} features all three benchmark datasets while Secs.~\ref{exp:attention} and \ref{exp:classification} only implements the first two; this decision followed the procedure adopted by the benchmark models considered, as our aim was to provide a sound comparison to the state of the art. All data were standardised per channel. 
	\\
		
	\noindent\textbf{Implementation details.} A unique AGG architecture was implemented for all datasets, meaning that no experiment-specific hyperparameter tuning was performed. AGG featured two encoder layers (Sec.~\ref{sec:graph_encoder}) and one conditional attention generator (Sec.~\ref{sec:conditional_attention}), followed by a regression or classification head depending on the task. All embeddings were 16 dimensions per feature with 8 attention heads. The MLPs in equations~\ref{eq:encoder_out}~\&~\ref{eq:cross_out} featured 2 layers: an input layer of dimension $5\times16=80$ and, following the design of \cite{vaswani2017attention, devlin2018bert}, a hidden layer of dimension equal to $\text{number of heads}\times\text{embedding dimension} = 8\times80=640$, which was then reduced back to the embedding dimension ($80$). During training, we used a Dropout rate of 0.2 for both the MLP layers and the attention masking. As a result, this implementation of AGG had 378k trainable parameters with a standard context length of $L=100$ nodes, which are padded if the context length exceeds the dataset such as some samples in the \emph{Physionet} dataset. We used ADAM~\cite{kingma2014adam} with an initial learning rate of 0.005 which was then annealed to 0.001 over the course of 200 epochs, gradient clipping $=1.0$ was also used.  Refer to Fig.~\ref{fig:architecture} for more details of the AGG architecture.
	\\
	
	\noindent\textbf{Infrastructure.} AGG was implemented on PyTorch \cite{pytorch} using an Nvidia RTX Titan GPU with 24GB of VRAM and 4608 CUDA Cores, and an Intel Core i9-9900K with 16 cores and 32GB of RAM running Ubuntu 22.04 64bit. Code is available\footnote{https://github.com/ChristopherLey/AsyncGraphGenerator}.

	\subsection{E1: Data imputation under increasing percentage of missing data }
	\label{exp:missing}
	Following \cite{miao2021generative}, we evaluated AGG's imputation performance wrt the rate of missing data. To this end, we randomly split the data into $r\%$ targets and $(100-r)\%$ inputs (see Figs.~\ref{fig:data_collection} and \ref{fig:architecture}), with the targets split again in $80\%-20\%$ for training and validation respectively. We chose $r\in\{10,30,50,70,90\}$ thus approaching an extremely-sparse imputation task. The Root Mean Square Error (RMSE) was considered as the evaluation metric. See Appendices \ref{app:kdd_data_prep} for details about the Beijing dataset and \ref{app:block_setup} for data removal and batching and Table~\ref{tab:rsme-imputation-performance-extended} for extended results.
	\\
	
	
	Table~\ref{tab:rsme-imputation-performance} shows the performance of the methods considered, alongside the baseline Mean (which just reports the mean of neighbouring nodes as prediction). Across all percentages of missing data ($r$), AGG outperformed all benchmarks and exhibited an average improvement  of 21.3\% on PhysioNet, 59.6\% on Beijing PM2.5 dataset, and 69.5\% on UCI (wrt SSGAN). A keen observer would note that, unlike previous methods, AGG does not decrease its performance monotonically with $r$, in fact, it even improves with $r$ in some cases (note the improvement of $r=30\%$ vs $r=10\%$ on the Beijing dataset in the extended Table~\ref{tab:rsme-imputation-performance-extended}). We attribute this behaviour to two main reasons. First, AGG's robustness the data's sparsity level, meaning that the model sees little change in the underlying signal with $r\leq50\%$. Second, AGG's sensitivity to data augmentation (see Sec.~\ref{sec:discuss}): it seems that $r=30\%$ is an inflection point whereby the loss of information (in an \emph{information theoretic} \cite{Shannon1949} sense) is too great to reconstruct the underlying signal dynamics accurately.
	\\

	\begin{table}
		\caption{Time series imputation performance (RMSE) for all models considered under different percentage of removed data ($r$).} 
		\label{tab:rsme-imputation-performance}
		\begin{adjustbox}{width=\columnwidth,center}
			\begin{tabular}{lccccccccc}
				\toprule
				Dataset &\multicolumn{3}{c}{UCI} &\multicolumn{3}{c}{PhysioNet} &\multicolumn{3}{c}{Air Quality}\\
				\cmidrule(r){1-10}
				Removed ($r$)	&10\%&50\%	&90\%&10\%&50\%&90\%&10\%&50\%	&90\% \\
				\midrule
				\multicolumn{10}{l}{Method}\\
				\cmidrule(r){1-1}
				GP-VAE \cite{fortuin2020gp} &0.670 &0.796 &0.882 &0.677 &0.787 &0.879 &0.522 &0.602 &0.771\\
				NAOMI \cite{liu2019naomi} &0.641 &0.794 &0.897 &0.632 &0.783 &0.865 &0.522 &0.602 &0.762\\
				BRITS \cite{cao2018brits} &0.621 &0.786 &0.867 &0.611 &0.779 &0.850 &0.531 &0.581 &0.720\\
				SSGAN \cite{miao2021generative}&0.600 &0.759 &0.841 &0.598 &0.762 &0.818 &0.435 &0.490 &0.660\\
				{\bf AGG} &{\bf 0.195} &{\bf 0.222} &{\bf 0.241} &{\bf 0.494} &{\bf 0.532} &{\bf 0.702} &{\bf 0.176} &{\bf 0.197} &{\bf 0.329}\\
				\bottomrule
			\end{tabular}
	\end{adjustbox}
	\end{table}

	\subsection{E2: comparison to other attention-based methods}
	\label{exp:attention}
	
	\begin{table}
		\small
		\begin{center}
			\caption{AGG vs state of the art on imputation of two datasets (MAE). Best scores denoted in bold font.} 
			\label{tab:attention_comparison}
			\begin{tabular}{ccccc}
				\toprule
				Method  &\multicolumn{3}{c}{Healthcare} 	&Air Quality\\
				\midrule
				Missing rate &$10\%$	&$50\%$	&$90\%$	&$13\%$\\
				\midrule
				GRIN \cite{cini2021filling}	&-		&-	&-	&12.08\\
				MTSIT \cite{yildiz2022multivariate}	&0.228	&0.236	&{\bf0.238}	&8.31 \\
				\midrule
				{\bf AGG} 	&{\bf0.220}	&{\bf0.229}	&0.240	&{\bf 7.32}\\
				\bottomrule
			\end{tabular}
		\end{center}
	\end{table}
	
	We considered the exact experimental methodology of both GRIN \cite{cini2021filling} (only Air Quality) and MTSIT \cite{yildiz2022multivariate}, who selectively removed samples from the Air Quality Index dataset following the original design of BRITS~\cite{cao2018brits}. See Appendix~\ref{app:kdd_data_prep} for more details. We considered this scheme in order to directly compare to state-of-the-art scores of methodologies that closely resemble that of the AGG, mainly a Graph Neural Network (GRIN) \cite{cini2021filling} and an attention based method, a Transformer (MTSIT) \cite{yildiz2022multivariate}. From Table~\ref{tab:attention_comparison}, we can see that the AGG has comparable results to MTSIT for the Healthcare (Physionet) dataset, which lacks topological information to be exploited and thus it is reasonable that both attention mechanisms behave similarly. Air Quality, on the other hand, has a strong topological component and we can see the reasonable improvement ($12\%$ and $34\%$ respectively) reported by AGG compared to MTSIT and GRIN.

	\subsection{E3: Classification and prediction}
	\label{exp:classification}
	
	\begin{table}
		\small
		\begin{center}
			\caption{Performance of pre-trained models on classification (PhysioNet ICU mortality) \& prediction (AQI Air Quality PM2.5). Mean and standard deviation values of 5-fold cross validation of AUC and MAE are shown.} 
			\label{tab:post-imputation}
			\begin{tabular}{ccc}
				\toprule
				Method  &PhysioNet (AUC) 	& Air Quality (MAE)\\
				\cmidrule(r){1-3}
				GRIN \cite{cini2021filling}	&-	&$10.23$\\
				BRITS \cite{cao2018brits}	&$0.850\pm 0.002$	&11.56 \\
				\midrule
				{\bf AGG} &$\mathbf{0.862\pm 0.007}$ &$\mathbf{3.64 \pm 0.005}$\\
				\bottomrule
			\end{tabular}
		\end{center}
	\end{table}
	Following the methodologies of \cite{cao2018brits,miao2021generative}, the model pretrained on the imputation task was used to predict in-hospital mortality on Physionet. Specifically, we fine-tuned the model pretrained with $10\%$ of data removed as explained above and, similarly to BRITS, we performed $k$-fold ($k=5$) cross validation with the entire dataset. AGG achieved an average $\text{\bf AUC}=\mathbf{0.862}$, thus improving over BRITS which reported $\text{AUC}=0.850$ \cite{cao2018brits}. Though SSGAN did not report an exact performance index for this experiment, from Fig.4a in \cite{miao2021generative} SSGAN appeared to perform on par with BRITS with $\text{AUC}\simeq 0.85$.
	\\
	
	AGG was then used to predict Air Quality (PM2.5, Beijing dataset) and compared against the two best-scoring methodologies encountered in the literature following the setting in \cite{yi2016st} regarding the test/train split and the use of MAE. In this task, AGG scored a prediction $\text{\bf MAE}=\mathbf{3.64}$ thus outperforming both BRITS \cite{cao2018brits} and GRIN \cite{cini2021filling} as showed in Table \ref{tab:post-imputation}. We conjecture that the considerable improvement of AGG ($64.4\%$) wrt GRIN can be explained by its ability to incorporate learnt spatial information (through the channel feature encoding), thus effectively learning an accurate phase shift among locations, compared to the hand crafted topological graph utilised in GRIN.

	\subsection{E4: Imputation of financial time series}\label{exp:forex}

	{In order to demonstrate the effectiveness of the AGG as a general framework for time series imputation beyond standard benchmarks, we included additional experimentation on foreign exchange (Forex) data under various regimes of removed data. A more comprehensive description of this experimental setting can be found in~\ref{app:forex_app}.}\\
	
	{Table \ref{tab:rsme-imputation-forex} reports the MAE and RMSE reported by AGG in this experiment. Notice that AGG performed well across all sparsity levels in terms of the considered performance metrics, in line with previous experimentation in Table \ref{tab:rsme-imputation-performance}. Critically, for a data removal rate of 30\%, AGG reports a MAE of only MAE = 0.043. A key point of interest is the improved performance of the 30\% sparsity vs the 10\%, this improvement is inline with previous performance gains in the Beijing Air Quality data (see Table~\ref{tab:rsme-imputation-performance-extended}), this is discussed in Section~\ref{exp:missing}, but briefly this is hypothesised to be a result of the simultaneous robustness to sparsity of the AGG and the increase in training samples available, a 3x increase. This improved performance due to increased augmentation in discussed in the next section, Section~\ref{sec:discuss}. \\
		
	These results also generally suggest two additional points, firstly; that the AGG performs well across a wide range of time-series domains and secondly; that foreign exchange data is strongly coupled across the various channels (currencies), an inductive bias that the AGG could seemingly exploit.}

	\begin{table}[!ht]
	\small
		\caption{Performance of AGG on the imputation of (daily) foreign exchange rate time series for varying levels of removed data ($r$). Mean Absolute Error (MAE) and Root Mean Square Error (RMSE) computed on test set. } 
		\label{tab:rsme-imputation-forex}
		\centering
			\begin{tabular}{lccccc}
				\toprule
				Dataset &\multicolumn{5}{c}{Forex} \\
				\cmidrule(r){1-6}
				Removed ($r$)	&10\%&30\%&50\%&70\%&90\% \\
				\midrule
				AGG (MAE) &0.047 &0.043 &0.049 &0.085 &0.111\\
				AGG (RMSE) &0.073 &0.071 &0.076 &0.128 &0.174\\
				\bottomrule
			\end{tabular}
	\end{table}

	\section{Discussion: On the effectiveness of data augmentation}
	\label{sec:discuss}
	
	\begin{figure}[!ht]
		\centering
		\includegraphics[width=0.80\textwidth]{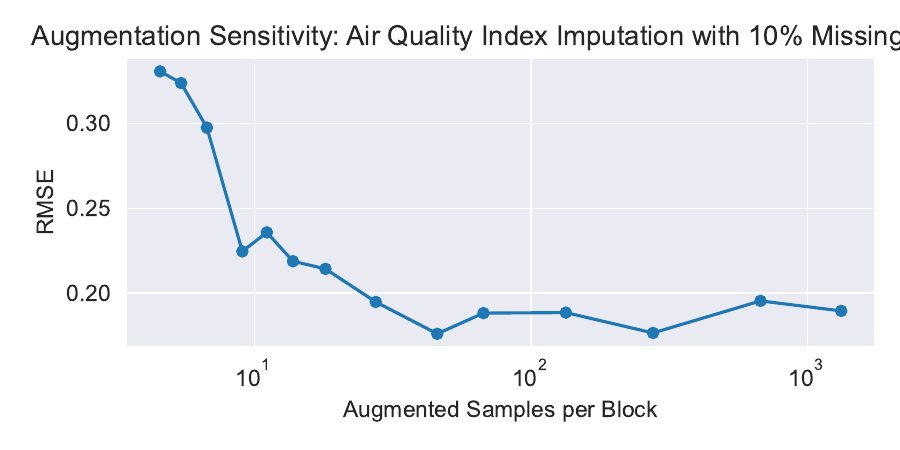}
		\caption{AGG performance vs number of training samples from the \emph{same} dataset via augmentation. }
		\label{fig:kdd10sizesensitivity}
	\end{figure}
	\noindent
	Distinguishing features of the AGG include invariance to sparsity (missing data) and ability to exploit translation equivariance. It is widely accepted that data augmentation regularises a model towards the applied transformations \cite{Balestriero2022,Neyshabur2017,Neyshabur2014}. If these transforms align with the geometric priors \cite{bronstein2021geometric} they can be exploited to create expressive representation of features in the signal space. This would allow the model to capture relevant interacting dynamics between channels, while ignoring superfluous information. It is expected that this inductive bias introduces some form of capacity control \cite{Neyshabur2017} which in turn allows for successful generalisation. 
	\\
	
	Data augmentation should then emphasise geometric priors in our model to fully learn a generalisable representation of the signal of interest. Our choice of augmentation is inspired by self supervised learning (SSL) \cite{misra2020self,zbontar2021barlow} in computer vision, where augmentations exploit the translation equivariance in images through shift operations. In the same vein, we randomly remove samples from the training set to promote sparsity in our dataset and shift the inputs (relative to targets) in order to leverage the translation equivariance. 
	\\
	
	We studied the effect of this approach to data augmentation on the imputation task with $10\%$ of the data removed (as defined in Sec.~\ref{sec:eval}). To this end, we varied the stride length of each sample: the finer the stride, the more data samples are generated from the same training data---more details in Appendices~\ref{app:block_setup} and \ref{app:sensitivity}. Fig.~\ref{fig:kdd10sizesensitivity} shows the effect of the number of augmented samples of each block on the imputation performance via RMSE over the validation set, as defined by \cite{yi2016st}. 
	\\
	
	The validation RMSE of AGG decreased sharply up to approximately 60x augmented samples, thus confirming the existence of a threshold for data augmentation in AGG after which complexity cost increases without gain in performance. This is consistent with \cite{Balestriero2022} who found empirically that 50x augmented samples were required to estimate their closed form of the loss. In general cases, this threshold should be determined based on the sampling theorem \cite{Shannon1949}, which relates the observation rate with the dynamic content of the signals (for the stationary case).

	\section{Conclusions}
	\noindent
	We have presented asynchronous graph generators (AGGs), a family of attention-based models for multichannel time series that represents observations as nodes of a dynamic graph without assuming temporal regularity or recurrence. Using data-augmentation techniques inspired from computer vision and learnable embeddings from language models, we have shown that AGG can be successfully trained under missing-data regimes to discover rich relationships among variables of interest. Once trained, AGG can be used for data imputation --and as a consequence classification and prediction-- by means of transductive node generation, that is, by generating a new node in the graph conditioned on timestamp and channel data. We achieve this by the novel concept of \emph{conditional attention generation}. We have experimentally validated the superiority of AGG against the state of the art on different relevant datasets and experimental conditions. Our simulations confirm the robustness of AGG to sparsity and sample asynchronicity, thus making it well suited for real-world applications involving incomplete multi-channel time-series data.
	
	\section*{Limitations}
	\noindent
	Dependencies learnt by AGG can be made arbitrarily complex by considering sufficient data and layers in the architecture; see Fig.~\ref{fig:architecture}. We addressed the effect of data augmentation in Sec.~\ref{sec:discuss}, but AGG is still lacking a methodology to choose a number of layers providing acceptable performance at reasonable computation. In our experiments, 2 encoder and 1 generator layers were enough for the benchmark datasets; however, massive complex datasets might require deeper AGG architectures. 
	{Lastly, there are known issues related to GNNs over-smoothing and over-squashing predictions or features \cite{zhang2021node,li-over-sq,giraldo_GNN} thus rendering them unreliable; these undesired properties might also be inherited by AGG and mitigating them  remains part of our future work.}

	\appendix
		\section{Appendix: Asynchronous Graph Generators}

		\subsection{Experimental Datasets}\label{app:datasets}
		In this section the reader can find more information regarding the three widely used benchmarks for multivariate time-series datasets that where used to compare the Asynchronous Graph Generator to other state-of-the-art works in Section~\ref{sec:eval}.
		\subsubsection{Beijing Air Quality Dataset}\label{app:kdd_data_prep}
		The air quality dataset, consists of PM2.5 measurements from 36 monitoring stations in Beijing. The measurements are hourly collected from 2014/05/01 to 2015/04/30. Overall, there are 13.3\% values are missing. For this dataset, we do pure imputation task with varying data removal and for the prediction task we do 6 hour prediction. Unlike the other datasets there is an explicit train/test split used in all prior works, which we followed as well in order to maintain comparable results \cite{cao2018brits,misc_beijing_pm2.5_data_381,luo2018multivariate,luo2019e2gan,silva2012predicting,yi2016st}. This involves removing the 3rd, 6th, 9th, and 12th month of the dataset as the test set, as outlined in \cite{yi2016st}, furthermore as approximately 13\% of the data is missing from the PM2.5 values of the remaining training set, if for example, a month proceeding the test months has any data missing, the same date and time is removed and used as the ground truth for the test imputation. The model is trained on a randomly removed data from the training set at the same rate of 13\%
		\subsubsection{Physionet 2012 ICU dataset}
		The ICU mortality prediction health-care data used in the PhysioNet Challenge 2012 \cite{silva2012predicting}, consists of 4000 multivariate clinical time series from intensive care unit (ICU). Each time series contains 35 measurements such as Albumin, heart-rate etc., which are irregularly sampled at the first 48 hours after the patient’s admission to ICU. We stress that this dataset is extremely sparse. There are up to 78\% missing values in total. We performed the imputation task with varying additional data removed as well as the post-imputation classification task.
		\subsubsection{UCI localization for Human Activity Dataset}
		The UCI localization data for human activity \cite{kaluvza2014multi} contains records of five people performing different activities such as walking, falling, sitting down etc (there are 11 activities). Each person wore four sensors on her/his left/right ankle, chest, and belt. Each sensor recorded a 3-dimensional coordinates for about 20 to 40 millisecond. The dataset was used for the imputation task as well as post-imputation activity classification.
		\subsubsection{Foreign Exchange Rate Data}\label{app:forex_app}
		{
		The foreign exchange rate data used in this experiment is the daily closing value of 22 currencies vs.~the United States 
		Dollar (USD). The currencies included are:		
		\begin{enumerate}
		\item AUSTRALIA - AUSTRALIAN DOLLAR/USD,
		\item EURO AREA - EURO/USD,
		\item NEW ZEALAND - NEW ZEALAND DOLLAR/USD,
		\item UNITED KINGDOM - UNITED KINGDOM POUND/USD,
		\item BRAZIL - REAL/USD,
		\item CANADA - CANADIAN DOLLAR/USD,
		\item CHINA - YUAN/USD,
		\item HONG KONG - HONG KONG DOLLAR/USD,
		\item INDIA - INDIAN RUPEE/USD,
		\item KOREA - WON/USD,
		\item MEXICO - MEXICAN PESO/USD,
		\item SOUTH AFRICA - RAND/USD,
		\item SINGAPORE - SINGAPORE DOLLAR/USD,
		\item DENMARK - DANISH KRONE/USD,
		\item JAPAN - YEN/USD,
		\item MALAYSIA - RINGGIT/USD,
		\item NORWAY - NORWEGIAN KRONE/USD,
		\item SWEDEN - KRONA/USD,
		\item SRI LANKA - SRI LANKAN RUPEE/USD,
		\item SWITZERLAND - FRANC/USD,
		\item TAIWAN - NEW TAIWAN DOLLAR/USD,
		\item THAILAND - BAHT/USD,
		\end{enumerate}
		\vspace{10pt}
		The dataset was obtained from \emph{Yahoo Finance}'s Application Programming Interface ( \url{https://developer.yahoo.com/api/}). The data spanned 20 years of daily financial data from 2000-01-03 to 2019-12-31, and included several major market events including the 2008 ``Global Financial Crisis'', so it is indicative of a wide spectrum of events in finance.
		\\
		
		The training ($60\%$), testing ($20\%$) and validation ($20\%$) dataset were selected randomly from 2-month blocks indexed by $i \in [3, 240]$, in proportions 3:1:1. Then, each 2 month block index ($i$) was converted to the relevant date range according to \begin{itemize}
			\item Year: $[2000 + \floor{\frac{i}{12}}, 2000 + \floor{\frac{i+1}{12}}]$, and 
			\item Month: $[i\pmod{12}, (i+1)\pmod{12}]$.
		\end{itemize} 
		All algorithms used to randomly create the test/train/validate split can be found in \url{https://github.com/ChristopherLey/AsyncGraphGenerator} as well as the exact masking used in the results found in Table~\ref{tab:rsme-imputation-forex}.

		}

		\subsection{Encoder Input Block Construction}\label{app:block_setup}
		In this section the reader can find more detailed instructions on how the data was deconstructed for data imputation as well as more detailed information regarding how the stride length affects the sample construction, which is directly related to data augmentation which is discussed in Section~\ref{sec:discuss}.
		
		\begin{figure}[!ht]
			\centering
			\includegraphics[width=1.0\linewidth]{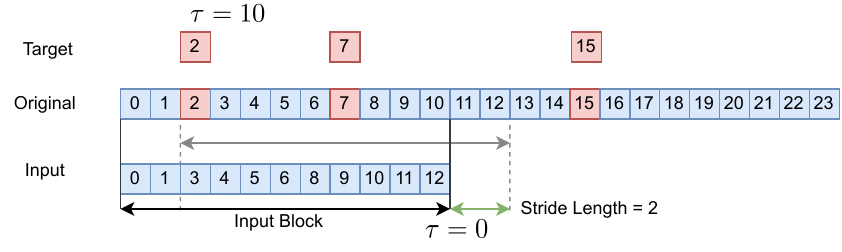}
			\caption{A pictographic representation of how the time series data is converted into an input block and a imputation target after data is randomly removed. The stride, as depicted in the image is defined as the number of steps the block is moved before it is considered the next input to the AGG. In this image the stride has value of 2 and input block a size of 11.}
			\label{fig:stridediagram}
		\end{figure}
		When building the imputation training dataset for the AGG we use the following procedure:
		\begin{enumerate}
			\item Randomly remove $r\%$ of the data, where $r \in \{10, 30, 50, 70, 90\}$
			\item For the inductive imputation (KDD) all removed data that falls in the validation range defined by \cite{yi2016st} is considered validation targets, the rest is considered training, similarly for the input data. For transductive imputation (Physionet and UCI datasets) randomly select from the removed data 20\% for the validation targets, the rest is considered the training targets.
			\item The remaining data that was not removed is considered as the inputs and is sorted temporally. 
			\item As depicted in Figure~\ref{fig:stridediagram}, given a predefined context length (input block length), if there is a target (e.g. value at $t=2$) within the range of the input block, $0 \leq t \leq 12$ then it is considered a valid target for imputation, relative time value $\tau$ is based on the largest value in the input block ($t=12 \implies \tau = 12 - t = 0$). The input along with that single target is considered one sample with $\tau = 12 - 2 = 10$. Samples are generated for the same input block for all targets in the input range e.g. $t =7$, where each target and input constitute an independent sample.
			\item Once all targets have been coupled with inputs the block is shifted by the stride length and the process is repeated until the end of the original input is reached.
		\end{enumerate}
		\subsection{Full Results}
		Extended results can be found in Table~\ref{tab:rsme-imputation-performance-extended}.
		\begin{table}[!ht]
				\caption{Time series imputation performance (RMSE) for all models considered under different percentage of removed data ($r$). Improvement denotes (as a percentage): AGG vs SSGAN.} 
				\label{tab:rsme-imputation-performance-extended}
				\begin{adjustbox}{width=\columnwidth, center}
				\begin{tabular}{lcccccccc}
					\toprule
					Dataset  &Removed ($r$) 	&Mean	&GP-VAE	\cite{fortuin2020gp}	&NAOMI \cite{liu2019naomi}		&BRITS \cite{cao2018brits}		&SSGAN \cite{miao2021generative}		&AGG	&Improvement\\
					\hline
					\multirow{ 5}{*}{UCI}	
					&10\%	&0.813	&0.670	&0.641	&0.621	&0.600 	&{\bf 0.195} &67.5\%\\
					&30\%	&0.873	&0.726	&0.724	&0.686	&0.666 	&{\bf 0.221} &66.8\%\\
					&50\%	&0.933	&0.796	&0.794	&0.786	&0.759 	&{\bf 0.222} &70.8\%\\
					&70\%	&0.943	&0.846	&0.854	&0.836	&0.803 	&{\bf 0.234} &70.9\%\\
					&90\%	&0.963	&0.882	&0.897	&0.867	&0.841 	&{\bf 0.241} &71.3\%\\
					\hline
					\multirow{ 5}{*}{PhysioNet}	
					&10\%	&0.799	&0.677	&0.632	&0.611	&0.598 	&{\bf 0.494}	&17.4\%\\
					&30\%	&0.863	&0.707	&0.703	&0.672	&0.670 	&{\bf 0.535}	&20.1\%\\
					&50\%	&0.916	&0.787	&0.783	&0.779	&0.762 	&{\bf 0.532}	&30.2\%\\
					&70\%	&0.936	&0.837	&0.835	&0.809	&0.782 	&{\bf 0.589}	&24.7\%\\
					&90\%	&0.952	&0.879	&0.865	&0.850	&0.818 	&{\bf 0.702}	&14.2\%\\
					\hline
					\multirow{ 5}{*}{Air Quality}		
					&10\%	&0.763	&0.522	&0.522	&0.531	&0.435 	&{\bf 0.176}	&59.5\%\\
					&30\%	&0.806	&0.562	&0.558	&0.561	&0.461 	&{\bf 0.157}	&65.9\%\\
					&50\%	&0.866	&0.602	&0.602	&0.581	&0.490 	&{\bf 0.197}	&59.8\%\\
					&70\%	&0.898	&0.709	&0.701	&0.641	&0.603 	&{\bf 0.225}	&62.7\%\\
					&90\%	&0.919	&0.771	&0.762	&0.720	&0.660 	&{\bf 0.329}	&50.2\%\\
					\bottomrule
				\end{tabular}
		\end{adjustbox}
		\end{table}
		\subsection{Sensitivity}\label{app:sensitivity}
\begin{figure}[ht!]
			\centering
			\begin{subfigure}[b]{\textwidth}
				\centering
				\includegraphics[width=0.8\textwidth]{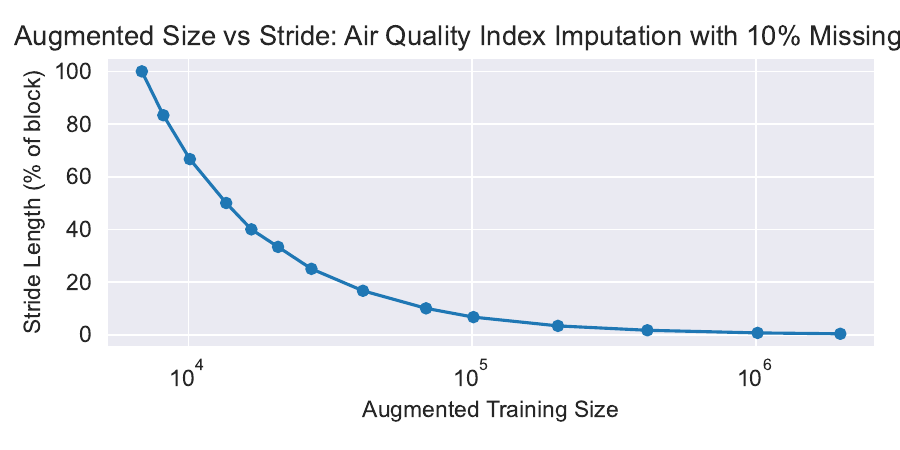}
				\caption{}
				\label{fig:kddsizevsstep}
			\end{subfigure}
			\begin{subfigure}[b]{\textwidth}
				\centering
				\includegraphics[width=0.8\textwidth]{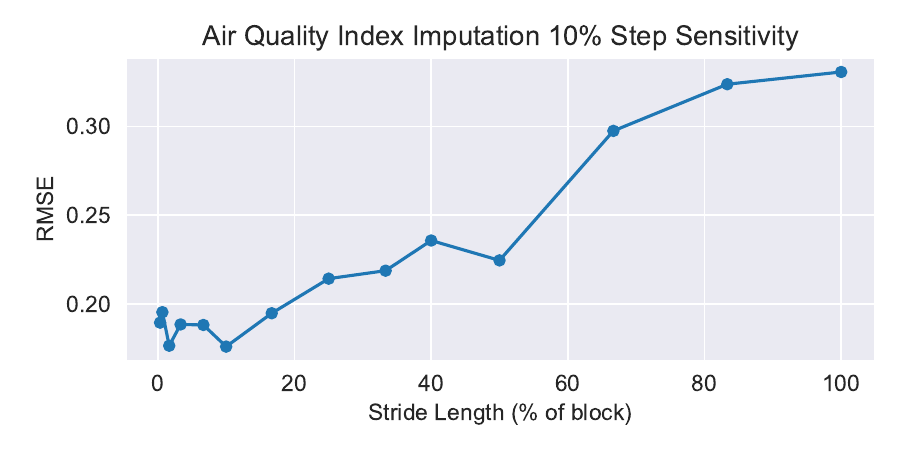}
				\caption{}
				\label{fig:kddstepsensitivity}
			\end{subfigure}
			\caption{(a) Augmented training size vs stride as a percentage of block size. (b) Stride sensitivity.}
		\end{figure}

		In this section the reader can find more information regarding why in the stride value elaborated in Appendix~\ref{app:block_setup} will directly affect the proportion of samples generated by the augmentation as seen in Figure~\ref{fig:kddsizevsstep}.
		
		We can see in Figure~\ref{fig:kddstepsensitivity}, that the smaller the stride value the better the performance because, as we can see in Figure~\ref{fig:kddsizevsstep}, the smaller the step the greater the training size. This large set of training examples will improve the performance up to limit, which is discussed in Section~\ref{sec:discuss} in more detail.

		\subsection{Prediction}
\begin{figure}[t]
			\centering
			\includegraphics[width=1.1\textwidth]{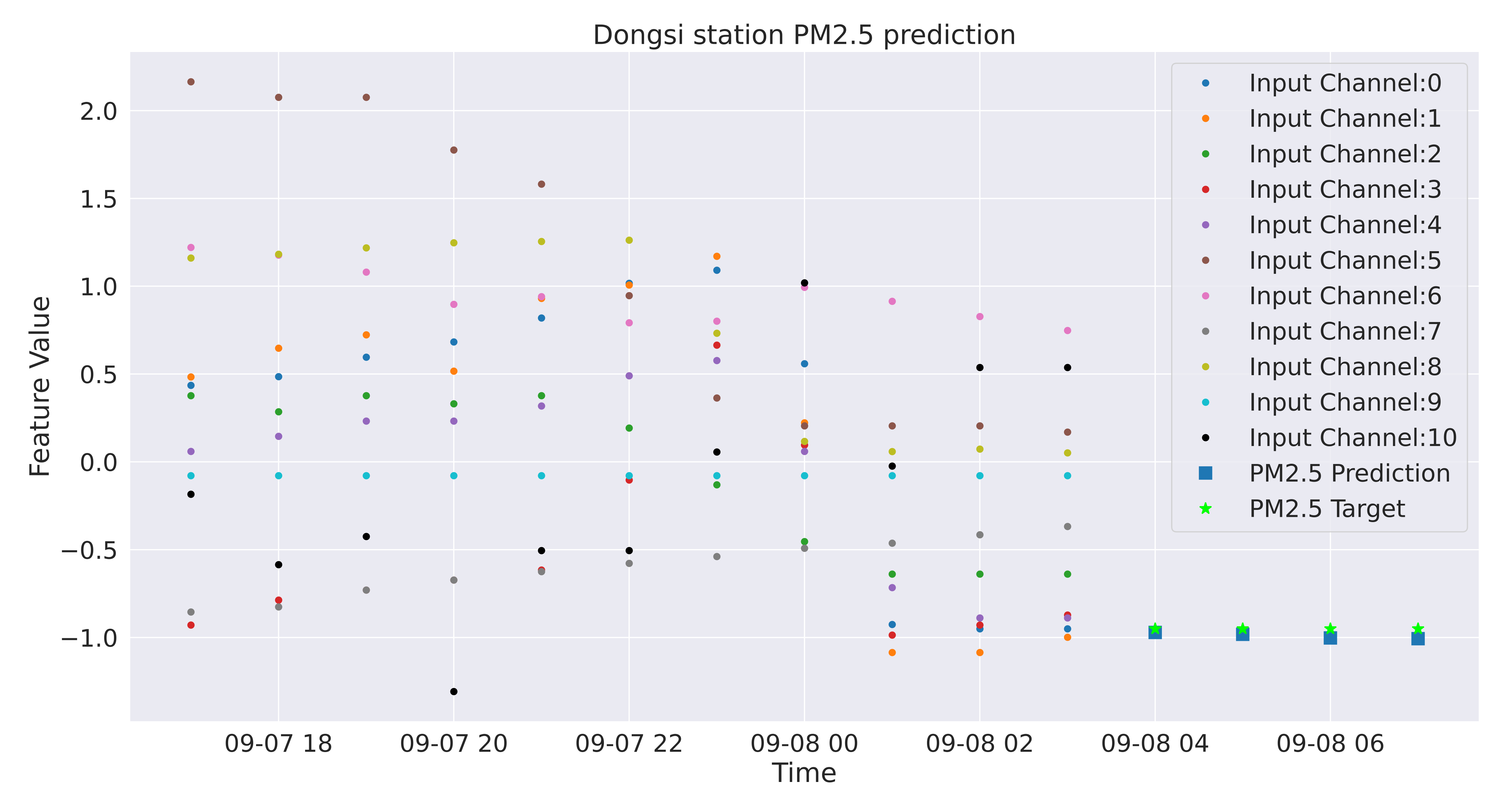}
			\caption{PM2.5 $\{1,2,3,4\}$-hour ahead PM2.5 prediction. Coloured dots indicate channel inputs for the given context window, squares indicate predicted values by AGG for various hours ahead, and stars indicate the ground-truth PM2.5 values. }
			\label{fig:predictionstationdongsi}
		\end{figure}

		We evaluated the prediction performance of the pre-trained model (30\% data removed imputation task) on the Beijing validation set. Table~\ref{tab:prediction_performance} shows the RMSE for the $n$-hour ahead prediction task with $n\in \{1,2,3,4\}$ for the entire validation set. All setting remains the same as in the imputation task in Sec.~\ref{sec:eval}. Figure~\ref{fig:predictionstationdongsi} shows the PM2.5 prediction for the particular case of the \emph{Dongsi} station, where dots indicate input channels, green stars are the hidden values of PM2.5 and the blue squares denote the predicted PM2.5 values. We emphasise that, as shown here, prediction is straightforwardly achieved by conditioning the generation on a \emph{future} time value that succeeds that of the input; furthermore, this required no modification or update of the trained AGG described in the main body of the paper.
		
		In the same line as the general imputation results shown in Sec.~\ref{sec:eval}, AGG as outperformed the benchmarks in this prediction setup too. Based on the results reported by \cite[Fig.~4]{miao2021generative}, AGG provides lower prediction error (RMSE) than SSGAN \cite{miao2021generative} , BRITS \cite{cao2018brits}, GP-VAE \cite{fortuin2020gp} and NAOMI \cite{liu2019naomi}.  
		\begin{table}
			\small
			\begin{center}
				\caption{Validation performance for PM2.5 (Air Quality Index dataset)} 
				\label{tab:prediction_performance}
				\begin{tabular}{|m{10em}|m{10em}|}
					\hline
					Prediction horizon (hours ahead)  &PM2.5 RMSE (over validation set)\\
					\hline
					1	&$0.387$\\
					2	&$0.409$\\
					3	&$0.455$\\
					4	&$0.453$\\
					\hline
				\end{tabular}
			\end{center}
		\end{table}
		
	\clearpage
	\bibliographystyle{elsarticle-num} 
	\bibliography{agg}

\begin{thebibliography}{10}
\expandafter\ifx\csname url\endcsname\relax
  \def\url#1{\texttt{#1}}\fi
\expandafter\ifx\csname urlprefix\endcsname\relax\def\urlprefix{URL }\fi
\expandafter\ifx\csname href\endcsname\relax
  \def\href#1#2{#2} \def\path#1{#1}\fi

\bibitem{miao2019answering}
X.~Miao, Y.~Gao, S.~Guo, L.~Chen, J.~Yin, Q.~Li, Answering skyline queries over
  incomplete data with crowdsourcing, IEEE Transactions on Knowledge and Data
  Engineering 33~(4) (2019) 1360--1374.

\bibitem{silva2012predicting}
I.~Silva, G.~Moody, D.~J. Scott, L.~A. Celi, R.~G. Mark, Predicting in-hospital
  mortality of {ICU} patients: The {Physionet/Computing in Cardiology}
  challenge, in: IEEE Computing in Cardiology, 2012, pp. 245--248.

\bibitem{little2019statistical}
R.~J. Little, D.~B. Rubin, Statistical analysis with missing data, Vol. 793,
  John Wiley \& Sons, 2019.

\bibitem{yoon2017multi}
J.~Yoon, W.~R. Zame, M.~van~der Schaar, Multi-directional recurrent neural
  networks: A novel method for estimating missing data, in: Time series
  workshop at ICML, 2017.

\bibitem{yi2016st}
X.~Yi, Y.~Zheng, J.~Zhang, T.~Li, {ST-MVL:} filling missing values in
  geo-sensory time series data, in: Proceedings of the 25th International Joint
  Conference on Artificial Intelligence, 2016.

\bibitem{cao2018brits}
W.~Cao, D.~Wang, J.~Li, H.~Zhou, L.~Li, Y.~Li, {BRITS:} bidirectional recurrent
  imputation for time series, in: Advances in Neural Information Processing
  Systems, Vol.~31, Curran Associates, Inc., 2018.

\bibitem{cini2021filling}
A.~Cini, I.~Marisca, C.~Alippi,
  \href{https://openreview.net/forum?id=kOu3-S3wJ7}{Filling the g\_ap\_s:
  Multivariate time series imputation by graph neural networks}, in:
  International Conference on Learning Representations, 2022.
\newline\urlprefix\url{https://openreview.net/forum?id=kOu3-S3wJ7}

\bibitem{yoon2018estimating}
J.~Yoon, W.~R. Zame, M.~van~der Schaar, Estimating missing data in temporal
  data streams using multi-directional recurrent neural networks, IEEE
  Transactions on Biomedical Engineering 66~(5) (2018) 1477--1490.

\bibitem{yoon2018gain}
J.~Yoon, J.~Jordon, M.~van~der Schaar, {GAIN}: Missing data imputation using
  generative adversarial nets, in: J.~Dy, A.~Krause (Eds.), Proceedings of the
  35th International Conference on Machine Learning, Vol.~80 of Proceedings of
  Machine Learning Research, PMLR, 2018, pp. 5689--5698.

\bibitem{liu2019naomi}
Y.~Liu, R.~Yu, S.~Zheng, E.~Zhan, Y.~Yue, {NAOMI:} non-autoregressive
  multiresolution sequence imputation, in: Advances in Neural Information
  Processing Systems, Vol.~32, Curran Associates, Inc., 2019.

\bibitem{yildiz2022multivariate}
A.~Y. Y{\i}ld{\i}z, E.~Ko{\c{c}}, A.~Ko{\c{c}}, Multivariate time series
  imputation with transformers, IEEE Signal Processing Letters 29 (2022)
  2517--2521.

\bibitem{bai2018empirical}
S.~Bai, J.~Z. Kolter, V.~Koltun, An empirical evaluation of generic
  convolutional and recurrent networks for sequence modeling, arXiv preprint
  arXiv:1803.01271 (2018).

\bibitem{chung2014empirical}
J.~Chung, C.~Gulcehre, K.~Cho, Y.~Bengio, Empirical evaluation of gated
  recurrent neural networks on sequence modeling, arXiv preprint
  arXiv:1412.3555 (2014).

\bibitem{lipton2016directly}
Z.~C. Lipton, D.~Kale, R.~Wetzel, Directly modeling missing data in sequences
  with rnns: Improved classification of clinical time series, in: Machine
  learning for healthcare conference, PMLR, 2016, pp. 253--270.

\bibitem{che2018recurrent}
Z.~Che, S.~Purushotham, K.~Cho, D.~Sontag, Y.~Liu, Recurrent neural networks
  for multivariate time series with missing values, Scientific reports 8~(1)
  (2018) 6085.

\bibitem{luo2018multivariate}
Y.~Luo, X.~Cai, Y.~Zhang, J.~Xu, Y.~xiaojie, Multivariate time series
  imputation with generative adversarial networks, in: Advances in Neural
  Information Processing Systems, Vol.~31, Curran Associates, Inc., 2018.

\bibitem{goodfellow2020generative}
I.~Goodfellow, J.~Pouget-Abadie, M.~Mirza, B.~Xu, D.~Warde-Farley, S.~Ozair,
  A.~Courville, Y.~Bengio, Generative adversarial networks, Communications of
  the ACM 63~(11) (2020) 139--144.

\bibitem{luo2019e2gan}
Y.~Luo, Y.~Zhang, X.~Cai, X.~Yuan, E2gan: End-to-end generative adversarial
  network for multivariate time series imputation, in: Proceedings of the 28th
  international joint conference on artificial intelligence, 2019, pp.
  3094--3100.

\bibitem{miao2021generative}
X.~Miao, Y.~Wu, J.~Wang, Y.~Gao, X.~Mao, J.~Yin, Generative semi-supervised
  learning for multivariate time series imputation, in: Proceedings of the AAAI
  conference on artificial intelligence, Vol.~35, 2021, pp. 8983--8991.

\bibitem{9046288}
Z.~Wu, S.~Pan, F.~Chen, G.~Long, C.~Zhang, P.~S. Yu, A comprehensive survey on
  graph neural networks, IEEE Transactions on Neural Networks and Learning
  Systems 32~(1) (2021) 4--24.

\bibitem{seo2018structured}
Y.~Seo, M.~Defferrard, P.~Vandergheynst, X.~Bresson, Structured sequence
  modeling with graph convolutional recurrent networks, in: Neural Information
  Processing, Springer International Publishing, 2018, pp. 362--373.

\bibitem{defferrard2016convolutional}
M.~Defferrard, X.~Bresson, P.~Vandergheynst, Convolutional neural networks on
  graphs with fast localized spectral filtering, in: Advances in Neural
  Information Processing Systems, Vol.~29, Curran Associates, Inc., 2016.

\bibitem{li2018diffusion}
Y.~Li, R.~Yu, C.~Shahabi, Y.~Liu,
  \href{https://openreview.net/forum?id=SJiHXGWAZ}{Diffusion convolutional
  recurrent neural network: Data-driven traffic forecasting}, in: International
  Conference on Learning Representations, 2018.
\newline\urlprefix\url{https://openreview.net/forum?id=SJiHXGWAZ}

\bibitem{atwood2016diffusion}
J.~Atwood, D.~Towsley, Diffusion-convolutional neural networks, in: Advances in
  Neural Information Processing Systems, Vol.~29, Curran Associates, Inc.,
  2016.

\bibitem{scarselli2008graph}
F.~Scarselli, M.~Gori, A.~C. Tsoi, M.~Hagenbuchner, G.~Monfardini, The graph
  neural network model, IEEE transactions on neural networks 20~(1) (2008)
  61--80.

\bibitem{li2015gated}
Y.~Li, R.~Zemel, M.~Brockschmidt, D.~Tarlow,
  \href{https://www.microsoft.com/en-us/research/publication/gated-graph-sequence-neural-networks/}{Gated
  graph sequence neural networks}, in: International Conference on Learning
  Representations, proceedings of iclr'16 Edition, 2016.
\newline\urlprefix\url{https://www.microsoft.com/en-us/research/publication/gated-graph-sequence-neural-networks/}

\bibitem{yu2017spatio}
B.~Yu, H.~Yin, Z.~Zhu, Spatio-temporal graph convolutional networks: A deep
  learning framework for traffic forecasting, Proceedings of the Twenty-Seventh
  International Joint Conference on Artificial Intelligence (IJCAI-18) (2017).

\bibitem{wu2019graph}
Z.~Wu, S.~Pan, G.~Long, J.~Jiang, C.~Zhang, Graph wavenet for deep
  spatial-temporal graph modeling, in: International Joint Conference on
  Artificial Intelligence, 2019, p. 1907–1913.

\bibitem{wu2020connecting}
Z.~Wu, S.~Pan, G.~Long, J.~Jiang, X.~Chang, C.~Zhang, Connecting the dots:
  Multivariate time series forecasting with graph neural networks, in:
  Proceedings of the 26th ACM SIGKDD International Conference on Knowledge
  Discovery \& Data Mining, 2020, p. 753–763.

\bibitem{vaswani2017attention}
A.~Vaswani, N.~Shazeer, N.~Parmar, J.~Uszkoreit, L.~Jones, A.~N. Gomez, L.~u.
  Kaiser, I.~Polosukhin, Attention is all you need, in: Advances in Neural
  Information Processing Systems, Vol.~30, Curran Associates, Inc., 2017.

\bibitem{cai2020traffic}
L.~Cai, K.~Janowicz, G.~Mai, B.~Yan, R.~Zhu, Traffic transformer: Capturing the
  continuity and periodicity of time series for traffic forecasting,
  Transactions in GIS 24~(3) (2020) 736--755.

\bibitem{zhang2018gaan}
J.~Zhang, X.~Shi, J.~Xie, H.~Ma, I.~King, D.-Y. Yeung, {GAAN:} gated attention
  networks for learning on large and spatiotemporal graphs, Proceedings for The
  Association for Uncertainty in Artificial Intelligence Conference (2018).

\bibitem{velivckovic2023everything}
P.~Veličković, \href{https://arxiv.org/pdf/2301.08210}{Everything is
  connected: Graph neural networks}, Current opinion in structural biology 79
  (2023) 102538.
\newblock \href {https://doi.org/10.1016/j.sbi.2023.102538}
  {\path{doi:10.1016/j.sbi.2023.102538}}.
\newline\urlprefix\url{https://arxiv.org/pdf/2301.08210}

\bibitem{brodyattentive}
S.~Brody, U.~Alon, E.~Yahav,
  \href{https://openreview.net/forum?id=F72ximsx7C1}{How attentive are graph
  attention networks?}, in: International Conference on Learning
  Representations, 2022.
\newline\urlprefix\url{https://openreview.net/forum?id=F72ximsx7C1}

\bibitem{velivckovicgraph}
P.~Veli{\v{c}}kovi{\'c}, G.~Cucurull, A.~Casanova, A.~Romero, P.~Li{\`o},
  Y.~Bengio, Graph attention networks, in: International Conference on Learning
  Representations, 2018.

\bibitem{rossi2020temporal}
E.~Rossi, B.~Chamberlain, F.~Frasca, D.~Eynard, F.~Monti, M.~Bronstein,
  Temporal graph networks for deep learning on dynamic graphs, arXiv preprint
  arXiv:2006.10637 (2020).

\bibitem{Balestriero2022}
R.~Balestriero, I.~Misra, Y.~LeCun, A data-augmentation is worth a thousand
  samples: Analytical moments and sampling-free training, in: Advances in
  Neural Information Processing Systems, Vol.~35, Curran Associates, Inc.,
  2022, pp. 19631--19644.

\bibitem{devlin2018bert}
J.~Devlin, M.-W. Chang, K.~Lee, K.~Toutanova,
  \href{https://aclanthology.org/N19-1423}{{BERT}: Pre-training of deep
  bidirectional transformers for language understanding}, in: Conference of the
  North {A}merican Chapter of the Association for Computational Linguistics:
  Human Language Technologies, 2019, pp. 4171--4186.
\newline\urlprefix\url{https://aclanthology.org/N19-1423}

\bibitem{kazemi2019time2vec}
S.~M. Kazemi, R.~Goel, S.~Eghbali, J.~Ramanan, J.~Sahota, S.~Thakur, S.~Wu,
  C.~Smyth, P.~Poupart, M.~Brubaker, Time2vec: Learning a vector representation
  of time, arXiv preprint arXiv:1907.05321 (2019).

\bibitem{bengio2000neural}
Y.~Bengio, R.~Ducharme, P.~Vincent,
  \href{https://proceedings.neurips.cc/paper_files/paper/2000/file/728f206c2a01bf572b5940d7d9a8fa4c-Paper.pdf}{A
  neural probabilistic language model}, in: Advances in Neural Information
  Processing Systems, Vol.~13, MIT Press, 2000.
\newline\urlprefix\url{https://proceedings.neurips.cc/paper_files/paper/2000/file/728f206c2a01bf572b5940d7d9a8fa4c-Paper.pdf}

\bibitem{bronstein2021geometric}
M.~M. Bronstein, J.~Bruna, T.~Cohen, P.~Veli{\v{c}}kovi{\'c}, Geometric deep
  learning: Grids, groups, graphs, geodesics, and gauges, arXiv preprint
  arXiv:2104.13478 (2021).

\bibitem{ba2016layer}
J.~L. Ba, J.~R. Kiros, G.~E. Hinton, Layer normalization, arXiv preprint
  arXiv:1607.06450 (2016).

\bibitem{maas2013rectifier}
A.~L. Maas, A.~Y. Hannun, A.~Y. Ng, et~al., Rectifier nonlinearities improve
  neural network acoustic models, in: Proceedings of the International
  Conference on Machine Learning, Vol.~30, Atlanta, GA, 2013, p.~3.

\bibitem{hinton2012improving}
G.~E. Hinton, N.~Srivastava, A.~Krizhevsky, I.~Sutskever, R.~R. Salakhutdinov,
  Improving neural networks by preventing co-adaptation of feature detectors,
  arXiv:1207.0580 (2012).
\newblock \href {http://arxiv.org/abs/1207.0580} {\path{arXiv:1207.0580}}.

\bibitem{he2016deep}
K.~He, X.~Zhang, S.~Ren, J.~Sun, Deep residual learning for image recognition,
  in: Proceedings of the IEEE conference on computer vision and pattern
  recognition, 2016, pp. 770--778.

\bibitem{granger1969investigating}
C.~W. Granger, Investigating causal relations by econometric models and
  cross-spectral methods, Econometrica: journal of the Econometric Society
  (1969) 424--438.

\bibitem{fortuin2020gp}
V.~Fortuin, D.~Baranchuk, G.~Raetsch, S.~Mandt, {GP-VAE:} deep probabilistic
  time series imputation, in: International Conference on Artificial
  Intelligence and Statistics, Vol. 108, 2020, pp. 1651--1661.

\bibitem{kaluvza2014multi}
B.~Kalu{\v{z}}a, B.~Cvetkovi{\'c}, E.~Dovgan, H.~Gjoreski, M.~Gams,
  M.~Lu{\v{s}}trek, V.~Mirchevska, A multi-agent care system to support
  independent living, International journal on artificial intelligence tools
  23~(01) (2014) 1440001.

\bibitem{kingma2014adam}
D.~P. Kingma, J.~Ba, Adam: A method for stochastic optimization, arXiv preprint
  arXiv:1412.6980 (2014).

\bibitem{pytorch}
A.~Paszke, S.~Gross, F.~Massa, A.~Lerer, J.~Bradbury, G.~Chanan, T.~Killeen,
  Z.~Lin, N.~Gimelshein, L.~Antiga, A.~Desmaison, A.~Kopf, E.~Yang, Z.~DeVito,
  M.~Raison, A.~Tejani, S.~Chilamkurthy, B.~Steiner, L.~Fang, J.~Bai,
  S.~Chintala, {PyTorch: An Imperative Style, High-Performance Deep Learning
  Library}, in: Advances in Neural Information Processing Systems 32, 2019, pp.
  8024--8035.

\bibitem{Shannon1949}
C.~Shannon, Communication in the presence of noise, Proceedings of the {IRE}
  37~(1) (1949) 10--21.

\bibitem{Neyshabur2017}
B.~Neyshabur, Implicit regularization in deep learning, arXiv preprint
  arXiv:1709.01953 (2017).

\bibitem{Neyshabur2014}
B.~Neyshabur, R.~Tomioka, N.~Srebro, In search of the real inductive bias: On
  the role of implicit regularization in deep learning, arXiv preprint
  arXiv:1412.6614 (2014).

\bibitem{misra2020self}
I.~Misra, L.~v.~d. Maaten, Self-supervised learning of pretext-invariant
  representations, in: Proceedings of the IEEE/CVF conference on computer
  vision and pattern recognition, 2020, pp. 6707--6717.

\bibitem{zbontar2021barlow}
J.~Zbontar, L.~Jing, I.~Misra, Y.~LeCun, S.~Deny, Barlow twins: Self-supervised
  learning via redundancy reduction, in: International Conference on Machine
  Learning, Vol. 139, 2021, pp. 12310--12320.

\bibitem{zhang2021node}
W.~Zhang, M.~Yang, Z.~Sheng, Y.~Li, W.~Ouyang, Y.~Tao, Z.~Yang, B.~CUI, Node
  dependent local smoothing for scalable graph learning, in: Advances in Neural
  Information Processing Systems, 2021.

\bibitem{li-over-sq}
H.~Li, C.~Li, J.~Zhang, Y.~Ouyang, W.~Rong, Addressing over-squashing in {GNNs}
  with graph rewiring and ordered neurons, in: IJCNN, 2024, pp. 1--8.

\bibitem{giraldo_GNN}
J.~H. Giraldo, K.~Skianis, T.~Bouwmans, F.~D. Malliaros, On the trade-off
  between over-smoothing and over-squashing in deep graph neural networks, in:
  International Conference on Information and Knowledge Management, CIKM '23,
  2023, p. 566–576.

\bibitem{misc_beijing_pm2.5_data_381}
S.~Chen, {Beijing PM2.5 Data}, UCI Machine Learning Repository, {DOI}:
  https://doi.org/10.24432/C5JS49 (2017).

\end{thebibliography}

	%
	%
	%
	%

	\vfill
	
\end{document}